\documentclass[10pt,twocolumn,letterpaper]{article}

\usepackage[pagenumbers]{cvpr} %

\usepackage{graphicx}
\usepackage{amsmath}
\usepackage{amssymb}
\usepackage{booktabs}
\usepackage{multirow}
\usepackage{wrapfig}
\usepackage{capt-of}

\usepackage[pagebackref,breaklinks,colorlinks]{hyperref}

\newcommand{\splitatcommas}[1]{%
  \begingroup
  \begingroup\lccode`~=`, \lowercase{\endgroup
    \edef~{\mathchar\the\mathcode`, \penalty0 \noexpand\hspace{0pt plus 1em}}%
  }\mathcode`,="8000 #1%
  \endgroup
}

\newcommand{\nop}[1]{}

\newcommand{\G}{\mathcal{G}}

\newcommand{\E}{\mathcal{E}}

\newcommand{\Z}{\mathbf{Z}}

\newcommand{\V}{\mathcal{V}}

\newcommand{\J}{\mathcal{J}}

\newcommand{\p}{\mathbf{p}}

\renewcommand{\u}{\mathbf{u}}

\renewcommand{\v}{\mathbf{v}}

\newcommand{\w}{\mathbf{w}}

\begin{document}

\title{Deep Learning Assisted Optimization for 3D Reconstruction\\from Single 2D Line Drawings}

\author{Jia Zheng\textsuperscript{1} \quad
Yifan Zhu\textsuperscript{2}$^*$ \quad
Kehan Wang\textsuperscript{3}$^*$ \quad
Qiang Zou\textsuperscript{4} \quad
Zihan Zhou\textsuperscript{1} \\
\textsuperscript{1}Manycore Tech Inc. \quad
\textsuperscript{2}Nanjing University of Aeronautics and Astronautics \\
\textsuperscript{3}University of California, Berkeley \quad
\textsuperscript{4}State Key Lab of CAD\&CG, Zhejiang University \\
{\tt\small \url{https://manycore-research.github.io/cstr}}
}

\twocolumn[{
\maketitle
\renewcommand\twocolumn[1][]{#1}
\centering
\vspace{-5mm}
\includegraphics[width=\linewidth]{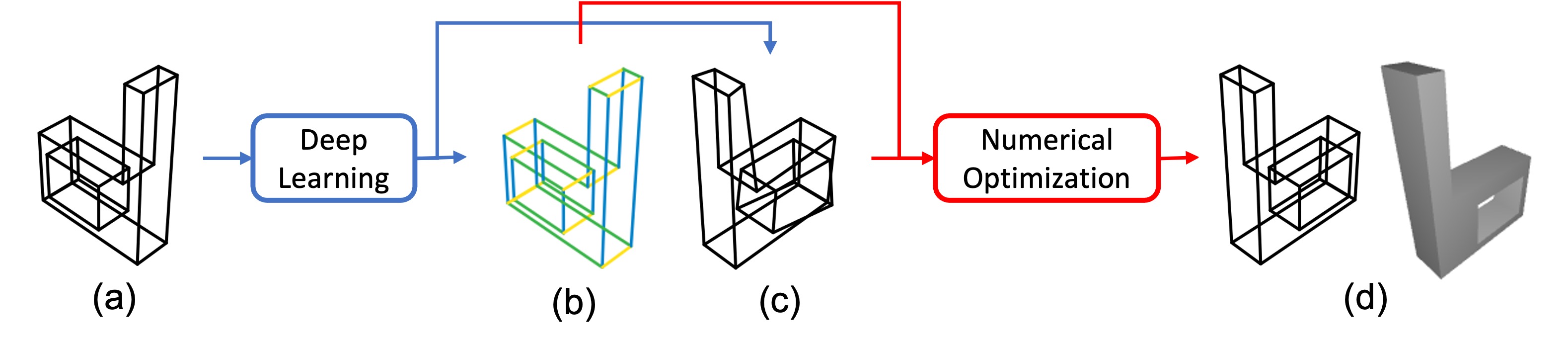}
\captionof{figure}{The overall pipeline. Given {\bf (a)} an input line drawing, we train deep models to predict {\bf (b)} geometric constraints (\eg, parallel constraints) and {\bf (c)} initial depth value of the vertices, which are then used in numerical optimization for geometric constraint solving to obtain an accurate and compact 3D model {\bf (d)}.\label{fig:pipeline}}
\vspace{3mm}
}]

\newcommand\blfootnote[1]{
\begingroup
\renewcommand\thefootnote{}\footnote{#1}
\addtocounter{footnote}{-1}
\endgroup
}
\blfootnote{$^*$Work done during internships at Manycore Tech Inc.}

\begin{abstract}
In this paper, we revisit the long-standing problem of automatic reconstruction of 3D objects from single line drawings. Previous optimization-based methods can generate compact and accurate 3D models, but their success rates depend heavily on the ability to (i) identifying a sufficient set of true geometric constraints, and (ii) choosing a good initial value for the numerical optimization. In view of these challenges, we propose to train deep neural networks to detect pairwise relationships among geometric entities (i.e., edges) in the 3D object, and to predict initial depth value of the vertices. Our experiments on a large dataset of CAD models show that, by leveraging deep learning in a geometric constraint solving pipeline, the success rate of optimization-based 3D reconstruction can be significantly improved.
\end{abstract}

\section{Introduction}
\label{sec:intro}

A line drawing is one of the simplest ways to represent a 3D object. It is widely adopted in computer-aided design (CAD) as a convenient means for designers to sketch ideas in the conceptual design stage~\cite{SheshC04,LeeFGG08,ChenKXDS08,ZouPLCFL15}. However, while the human vision system can interpret 2D line drawings as 3D objects with almost no difficulty, automatic 3D object reconstruction from single line drawings by the computer still remains a challenging problem~\cite{BonniciACCFFHIL19}.

The earliest work on the interpretation of line drawings of 3D objects treats it as a line labeling problem through which the realizability of a line drawing can be tested~\cite{Huffman71,Clowes71,Sugihara82}. Such a method is, however, too sensitive to modeling errors to be useful in practice. More recently, with the development of geometric constraint solving in CAD~\cite{BettigH11}, optimization-based method has become the most popular approach~\cite{Marill91,LeclercF92,LipsonS96,Varley03,SheshC04,WangCLT09}. In a typical optimization-based method, the variables are the missing depth values of the vertices in a line drawing. Various relationships between geometric entities \emph{in the 3D space}, such as parallel and perpendicular lines, are then identified. With enough constraints, a 3D model can be recovered by finding the values of these variables that minimize a certain objective function derived from those constraints.

In practice, however, an optimization-based method may fail to recover the desired shape due to the following major challenges. \emph{First}, detecting the geometric constraints is not a trivial task, as most relationships are not projection-invariant. For example, two perpendicular lines in 3D space can project to lines that form arbitrary angles in the 2D drawing. Thus, from a single line drawing, it is difficult to detect perpendicular lines in the 3D space. While a number of approaches may be employed for this purpose, there are often errors in the detection results. Including these errors in the numerical optimization will result in incorrect solutions. \emph{Second}, to solve the nonlinear optimization via a numerical method, an initial value of the variables is required. If the initial value is not properly chosen, the optimization may get trapped into local minima or fail to find the correct solution.\footnote{A nonlinear geometric constraint system has, in general, exponentially many solutions, and only one of them satisfies the reconstruction requirement.}

This work focuses on improving the performance of optimization-based 3D reconstruction by training deep neural networks to (i) detect geometric constraints in the line drawing, and (ii) predict initial values for the optimization (Figure~\ref{fig:pipeline}). Because of the varying number of vertices and edges in each line drawing, we adopt Transformer-based architectures~\cite{VaswaniSPUJGKP17} for both tasks. Specifically, for constraint detection, we treat each edge as a ``query'' edge and train a variant of Pointer Net~\cite{VinyalsFJ15} to predict one edge which forms a certain relationship with the query edge at one timestamp. For initial value prediction, we train a Transformer decoder to estimate the depth values of all vertices simultaneously. As  will be shown, our constraint detection method is better than traditional methods in terms of both accuracy and applicability. Furthermore, by leveraging deep learning for initial value prediction and constraint detection, the success rate of geometric constraint solving for 3D reconstruction is seen to be improved significantly. To the best of the authors' knowledge, integrating deep learning with nonlinear optimization for 3D reconstruction of line drawings has not been previously reported.

\section{Related Work}

\subsection{Optimization-based 3D Reconstruction}

There is a long line of work on optimization-based 3D reconstruction from 2D line drawings. Early work focuses on exploring various structural regularities and topological cues for the task. Marill~\cite{Marill91} proposed the MSDA principle, which minimizes the standard deviation of angles in the reconstructed object. 
Leclerc and Fischler~\cite{LeclercF92} combined MSDA with the face planarity constraint to handle more complex objects. Subsequent work~\cite{LipsonS96,Varley03,SheshC04} added more constraints including line parallelism, line verticality, isometry, corner orthogonality, skewed facial orthogonality, skewed facial symmetry, and so on. Wang~\etal~\cite{WangCLT09} developed new regularities for curved objects. However, little attention was paid to the issues of initialization and constraint applicability in these studies.

\smallskip
\noindent {\bf Initialization.} Most studies initialize the optimization from a flat 2D projection (\ie, setting the depths of all vertices to a fixed value)~\cite{Marill91,LipsonS96} or use randomized initialization for multiple times~\cite{LeclercF92}. Without available 3D information as guidance, the nonlinear optimization may not find the desired solution. To alleviate this issue, two possible strategies have been adopted in the literature.

The first strategy leverages knowledge about object topology or structural regularities to reduce the search space. Liu~\etal~\cite{LiuCLT08} used parameters of the planar faces of an object as variables of the objective function. This leads to a lower dimensional space where the optimization problem becomes easier to solve. Following this idea, Liu~\etal~\cite{LiuCT11} and Zou~\etal~\cite{ZouCFL15} proposed to decompose the line drawing of a complex object into multiple simpler ones, reconstruct 3D shapes from these simpler line drawings, and finally assemble the shapes into a complete object. But these methods require that true faces of the object (and a part decomposition for~\cite{LiuCT11,ZouCFL15}) to be provided, and can only handle polyhedral objects. Similarly, Tian~\etal~\cite{TianML09} used parallel lines to constrain the steps taken by an optimization method, so that line parallelism is preserved.

The second strategy tries to get an initial 3D model using non-optimization methods for certain types of objects~\cite{CompanyCCV04}, such as normalons (\ie, objects whose edges are all aligned with one of the three main axes). However, an initialization method for general objects remains elusive.

\smallskip
\noindent {\bf Constraint applicability.} Most work uses simple rules or heuristics to identify regularities in the 2D edge-vertex graph. This faces at least two major challenges in practice. \emph{First}, many regularities are not projection-invariant, therefore cannot be easily detected. 
\emph{Second}, error tolerance needs to be considered to deal with inaccuracies in the drawings if, say, the input drawings are derived from hand-drawn sketches or scanned images.

To improve the robustness of the interpretation, Lipson and Shapitalni~\cite{LipsonS96} introduced a continuous compliance function to indicate the probability that a detected regularity indeed represents a 3D geometrical relationship. The problem of constraint applicability was studied in~\cite{XuCSBMS14,GryaditskayaHLS20}. Instead of using all regularities at once, Xu~\etal~\cite{XuCSBMS14} used an initial baseline reconstruction and iterative scheme to identify applicable constraints. Similarly, Gryaditskaya~\etal~\cite{GryaditskayaHLS20} iteratively built a 3D scaffold by lifting one line at a time while using the current model to detect line intersections. However, these methods rely heavily on additional inputs in raw sketches, such as smooth-crossings, construction lines, and user drawing orders, which are not always available. We do not assume access to such special information in this paper.

More general approaches to identify redundant or inconsistent constraints were studied in~\cite{LangbeinMM04,ZouL07,ZouF20}. We use a similar scheme in our pipeline. Note that these methods can ensure the solvability of a geometric constraint system, but cannot distinguish correct constraints from incorrect ones.

\subsection{Deep Learning for Computer-Aided Design}

Recently, several generative models for parametric 2D sketches and 3D CAD models have been developed~\cite{SeffOZA20,WillisJLCP21,WuXZ21,GaninBLKS21,SeffZRA22}. At the core of these methods is a deep network (\eg, variants of Transformer) which predicts geometric entities and their relationships in a sequential manner. Our method shares similarity with these work in terms of network design. However, while these methods focus on modeling constraints in the 2D space, we aim to detect geometric constraints in 3D objects through their 2D projections.

Deep networks have also been extensively used to reconstruct 3D models from 2D sketches~\cite{NishidaGABB16,HuangKYM17,DelanoyAIEB18,LiPLTSW18,HanMLZ20,WangLYLCY20,LiPBM20,ZhangGG21,GuillardRYF21}. But most methods generate unstructured and noisy point clouds or meshes in an end-to-end fashion~\cite{DelanoyAIEB18,LiPLTSW18,HanMLZ20,WangLYLCY20,ZhangGG21,GuillardRYF21}. Nishida~\etal~\cite{NishidaGABB16} used deep learning to find the shape that best matches the input sketch among a set of pre-defined snippets. In~\cite{HuangKYM17,HanMLZ20}, the authors trained deep networks to predict executable programs in a domain specific language (DSL). Our method does not rely on any pre-defined primitives or DSLs, but aims to reconstruct compact (\ie, in the B-rep format) and accurate (\ie, within a normalized distance of $10^{-3}$) models using a general optimization framework.

\section{Problem Statement}

\begin{figure}[h]
    \setlength{\tabcolsep}{0.2pt}
    \begin{tabular}{c|c}
        \includegraphics[width=0.43\linewidth]{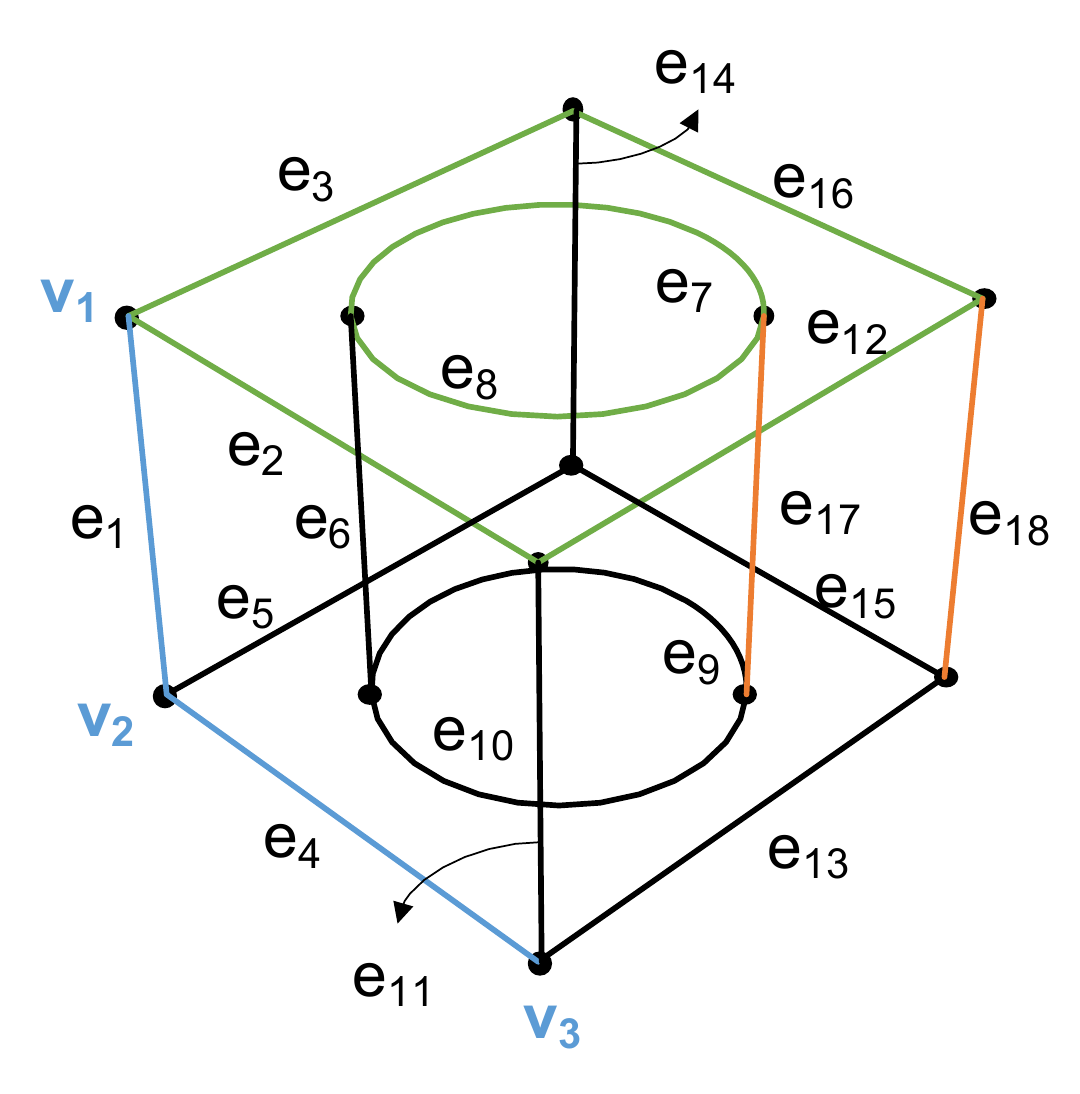} &
        \includegraphics[width=0.55\linewidth]{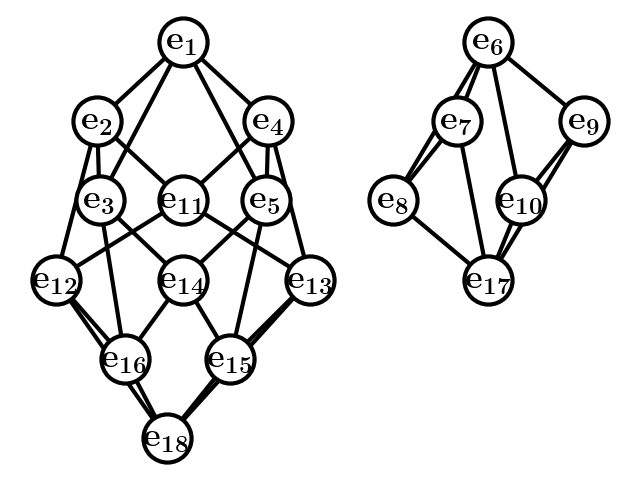} \\
    \end{tabular}
    \caption{{\bf Left}: An input line drawing. We also show examples of parallel (orange), perpendicular (blue), and face planarity (green) constraints. {\bf Right}: edge-vertex graph.}
    \label{fig:wireframe}
\end{figure}

In this paper, a line drawing is assumed to be a perspective projection\footnote{All techniques developed in this work are also applicable to orthographic projection, a special case where the camera is placed at infinity.} of a manifold object\footnote{We restrict our attention to manifold objects as CAD is primarily concerned with objects that can be physically manufactured.} in a generic view, so that all the edges (including silhouettes) and vertices of the object are visible. We further make two assumptions about the input: \emph{First}, the hidden lines and vertices are given. \emph{Second}, the crossing point of two edges is not a vertex. As illustrated in Figure~\ref{fig:wireframe}, the input line drawing can be represented as an edge-vertex graph $\G = (\V, \E)$ where each edge (or vertex) of the graph corresponds to exactly one edge (or vertex) of the object. Note that the graph may contain one or more connected components.

Our goal is to reconstruct a 3D model of the object. According to the pinhole camera model, for a 3D vertex $V=(X,Y,Z)$, its projection $\v = (x,y)$ on the image plane can be written as: $x = f\frac{X}{Z}, y = f\frac{Y}{Z}$, where $f$ is the focal length of the camera. Therefore, given the line drawing $\G = (\V, \E)$ with $m$ vertices, where each 2D vertex $\v_i\in \V$ is given by $\v_i = (x_i, y_i)$, our problem becomes estimating the depth values $\{Z_i\}_{i=1}^m$ for all the vertices.

\subsection{Geometric Constraint System}
\label{sec:geometric_constraint_system}

A geometric constraint system contains a set of constraints associating the variables (the depths $\{Z_i\}_{i=1}^m$ in our case), which allow these variables to be determined. For example, as shown in Figure~\ref{fig:wireframe}, a perpendicular constraint (in the 3D space) between edges $e_1$ and $e_4$ can be written as a nonlinear equation of the associated vertices ($V_1, V_2, V_3$):
\begin{align*}
    & \left( \frac{x_1 x_2 + y_1 y_2}{f^2} + 1 \right) Z_1 Z_2 - \left( \frac{x_1 x_3 + y_1 y_3}{f^2} + 1 \right) Z_1 Z_3 \\ 
    & - \left( \frac{x_2^2 + y_2^2}{f^2} + 1 \right) Z_2^2 + \left( \frac{x_2 x_3 + y_2 y_3}{f^2} + 1 \right) Z_2 Z_3 = 0. 
\end{align*}

More than a dozen types of constraints have been proposed in the literature. When enough constraints are found, a 2D line drawing can be lifted to 3D via geometric constraint solving. 

However, because many 3D relationships are not projection-invariant, detecting them from 2D line drawings is a non-trivial task. For example, under the pinhole camera model, parallel lines in 3D space project to converging lines in the 2D drawing~\cite{HartleyR2000}. Thus, vanishing point (VP) detection is often used to detect \emph{parallel constraints}. But 2D lines may also intersect at a common point by accident, resulting in false detections. As another example, \emph{face planarity constraints} are widely used in the literature~\cite{CompanyCCV04}. But detecting faces in a line drawing itself is a highly complex problem~\cite{MarkowskyW80,Hanrahan82,DuttonB83,Ganter83,BrewerC86,BagaliW95,ShpitalniL96,LiuLC02,LiuT04,VarleyC10,FangLL15}. Most existing methods are restricted to polyhedral objects only, and cannot guarantee successful recovery except for certain special types of objects (namely, objects with genus 0 and whose projections are 3-connected). 

To summarize, a constraint detection method may produce incorrect constraints. If such constraints are included in the numerical optimization, an incorrect 3D model will be obtained. At the same time, the number of detected constraints could exceed the number of variables. In other words, there may be \emph{redundant} and \emph{inconsistent} constraints in the system. In the following, we present a method for selecting a sufficient and consistent set of constraints.

\begin{figure*}[t]
    \centering
    \includegraphics[width=0.95\linewidth]{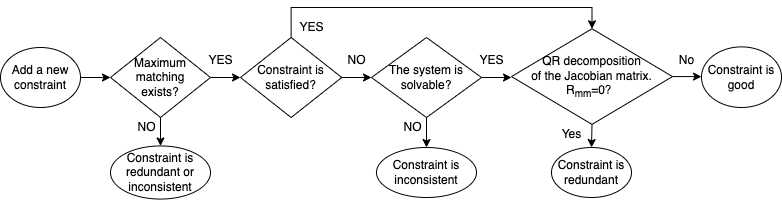}
    \caption{Flow chart for selecting a consistent and non-redundant constraint.}
    \label{fig:flowchart}
    \vspace{-3mm}
\end{figure*}

\subsection{Selecting and Solving the Constraints}
\label{sec:selecting_constraint}

Let $f_k(\Z)=0, k = 1, 2, \ldots, K$ be a set of $K$ nonlinear equations derived from the detected constraints, where $\Z = \{Z_i\}_{i=1}^m$ denote the depth variables. When the system is well-constrained, a solution for $\Z$ can be found using, for example, a least-square method: $\min \sum_{k=1}^K \left(f_k(\Z)\right)^2$.

But if some of the detected constraints are incorrect, the resulting equations may be \emph{inconsistent}. As a simple example, two constraints expressed by $Z_1 + Z_2 = 1$ and $Z_1 + Z_2 = -1$ are inconsistent. Including both constraints in the system would make it unsolvable. Note that more constraints than the number of variables, $K > m$, are usually generated in constraint detection. Therefore, we need to avoid the problem by selecting only $m$ consistent equations out of all detected constraints.

However, randomly choosing $m$ equations is unlikely to yield the desired result. This is because there may be redundancies in the system, which include \emph{structural redundancies} and \emph{numerical redundancies}. A structural redundancy over-constrains (a part of) the system. For example, there is a structural redundancy in the set of equations $f_k(Z_1, Z_2)=0, k = \{1, 2, 3\}$ because there are three equations but only two variables. A numerical redundancy occurs when one equation can be deduced from other equations in the system. For example, two constraints $Z_1 + Z_2 = 1$ and $2Z_1 + 2Z_2 = 2$ are numerically redundant because the second equation can be deduced from the first one.

\smallskip
\noindent {\bf Selecting the constraints.} We use a variant of the method proposed in~\cite{ZouL07} to iteratively select a sufficient set of $m$ equations. The procedure for selecting one consistent and non-redundant constraint is shown in Figure~\ref{fig:flowchart}. Below we briefly describe each key component. Readers are referred to~\cite{ZouL07} for more details.

As shown in Figure~\ref{fig:flowchart}, for each constraint added to the system, we first check for \emph{structural redundancy} by building a bipartite graph which has one node per variable, one vertex per equation, and one edge between a variable and an equation if the variable appears in the equation. The newly added constraint is not structurally redundant if a maximum matching of the graph can be found~\cite{Ait-AoudiaJM14}. Next, we check for \emph{numerical inconsistency} by solving the current system using a least-square method. The constraints are consistent if a solution can be found. Finally, we check for \emph{numerical redundancy} via QR decomposition of the associated Jacobian matrix. If the newly added constraint is numerically redundant, the last row in the matrix can be expressed as a linear combination of the other rows~\cite{ZouF19}. Consequently, the last entry in the $R$ matrix will be 0.

Note that, while the above procedure ensures the system is solvable, it does not guarantee that every selected constraint is correct. An incorrect constraint may be included as long as it is consistent and non-redundant w.r.t. all the previously selected constraints. 

\smallskip
\noindent {\bf Solving the constraints.} To solve the optimization problem, we adopt the \texttt{scipy.optimize.fsolve} API~\cite{fsolve}, which is a wrapper around MINPACK's \texttt{hybrd} and \texttt{hybrj} algorithms. Like all numerical methods for nonlinear optimization, an initial value of $\Z$ is needed. And it is known that the numerical procedure may fail to find a solution depending on the initial value as well as the system of equations. To improve its stability, we follow the common practice to repeat the process of selecting and solving the constraints for $N$ times, and choose the solution that satisfies the most constraints as the final output.

\smallskip
To summarize, we have now described the complete pipeline for reconstructing 3D model from 2D line drawing via geometric constraint solving. But to obtain the \emph{correct} 3D model, two issues remain:
\begin{enumerate}
    \item \emph{How to find as many true constraints as possible while minimizing the number of incorrect detections?}
    \item \emph{ How to choose the initial value of $\Z$?}
\end{enumerate}
In Sections~\ref{sec:constraint} and~\ref{sec:depth}, we present our learning-based methods to tackle these two problems, respectively. %

\section{Constraint Detection}
\label{sec:constraint}

Given the input line drawing $\G = (\V, \E)$, we observe that many constraints can be expressed as pairwise relationships between geometric entities (in the 3D space). For example, both parallel and perpendicular constraints are relationships between two edges. Therefore, in this section we develop a unified approach to detect pairwise relationships between geometric entities. Besides parallel and perpendicular constraints, other common pairwise relationships are also included: ``tangent'', ``equal length'', and ``symmetric'' constraints between two edges. 

Other constraints (\eg, face planarity) require the recovery of additional topological information (\ie, faces), which itself is a challenging problem. In this work, we adopt our previous work faceformer~\cite{WangZZ22} for face identification. Note that, if faces are known, pairwise constraints between the faces may also be detected using the techniques developed in this section.

\subsection{Constraint Detection Model}

Let $\E = \{ e_1, \ldots, e_n \}$ denote the set of all edges. For any given edge $e_{j_1}$, the number of edges satisfying a certain relationship (\eg, parallelism) with it may vary from 0 to $n-1$. We can write it as a sequence $g(e_{j_1}) = (e_{j_1}, e_{j_2}, \ldots, e_{j_s})$, where $e_{j_1}$ is regarded as the ``query'' edge and $e_{j_2}, \ldots, e_{j_s}$ are the $s-1$ edges related to it, with $0<s\leq n$.

This motivates us to formulate constraint detection as a sequence generation problem. And to detect geometric constraints among all edges, we may use every edge in $\E$ as the query edge and repeat the process for $n$ times. In the following, we describe how to generate the sequence $g(e_{j_1})$ given a query edge $e_{j_1}$.

\smallskip
\noindent{\bf Pointer Net.} To select a subset of the input edges that satisfy the geometric constraints with the query edge, we use Pointer Net~\cite{VinyalsFJ15}, a sequence-to-sequence model that learns a distribution over the input elements. Given the input sequence $P = \{\p_1, \p_2, \ldots\}$, it uses a deep network to learn the conditional probability $p(\J \mid P)$, where $\J = (j_1, \ldots, j_T)$ is a sequence of $T$ indices, each between $1$ and $|P|$: $p(\J \mid P) = \prod_{t=1}^T p (j_t \mid j_1, \ldots, j_{t-1}, P)$.

\begin{figure}[t]
    \centering
    \includegraphics[width=0.95\linewidth]{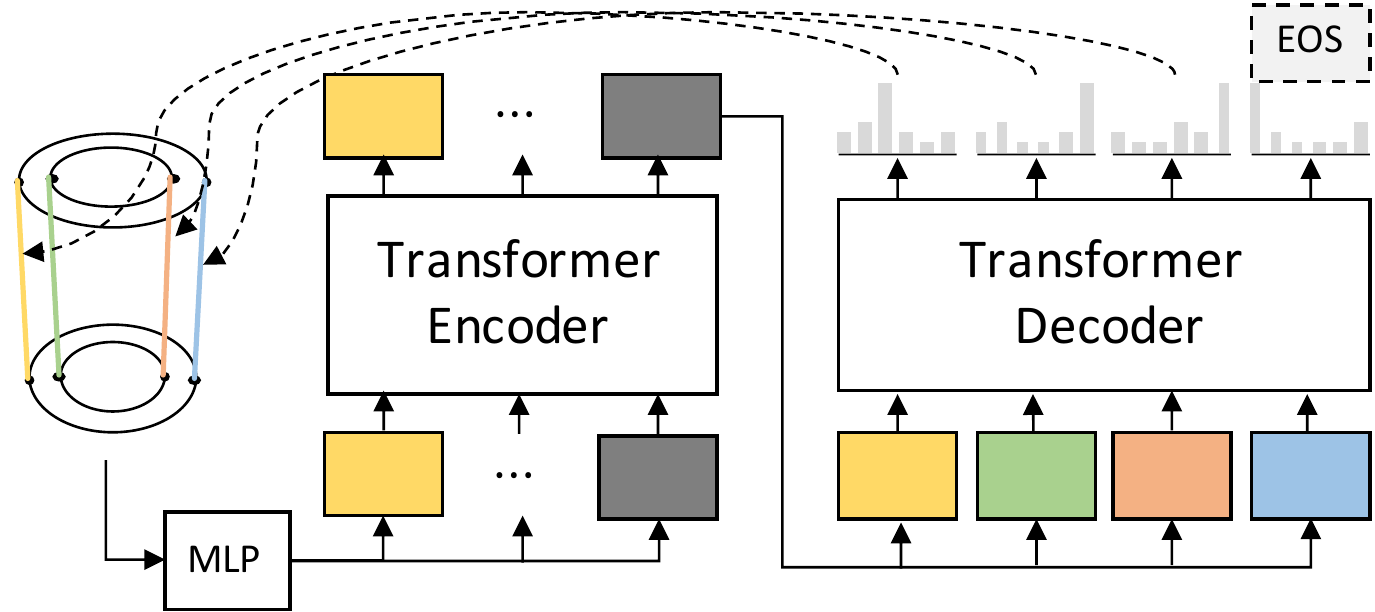}
    \caption{Constraint detection network.}
    \label{fig:constraint}
    \vspace{-5mm}
\end{figure}

The overall architecture is shown in Figure~\ref{fig:constraint}. First, an encoder is used to obtain a contextual embedding $\w_k$ for each input vector. Then, at each decoding time step $t$, the decoder outputs a pointer vector $\u_t$. We compare $\u_t$ with the contextual embeddings $\w_k$ via dot-product. Finally, a softmax layer is applied to produce a valid distribution over the input set $P$:
\begin{align}
    \{\w_k\}_{k=1}^{|P|} & = \mathbf{Encoder}(P; \theta) \\
    \u_t & = \mathbf{Decoder}(\J_{<t}, P; \theta) \\
    p(j_t = k \mid \J_{<t}, P; \theta) & = \mathbf{softmax}_{k}(\u_t^T \w_k)
\end{align}

In our model, both the encoder and decoder consist of six layers of standard Transformer blocks~\cite{VaswaniSPUJGKP17}. Each block has eight attention heads and a feed-forward dimension of 1024. We optimize the model parameters $\theta$ by maximizing conditional probabilities on the training set.

\smallskip
\noindent {\bf Input sequence and embeddings.} The input sequence consists of all edges $\E$. For each edge, we use a value embedding to indicate coordinate values of the edge, and a position embedding to indicate the edge location in the sequence. To deal with varying edge length, we uniformly sample a fixed number of edge points, and order the points lexicographically (from lowest to highest first by its $x$-coordinate, then by $y$-coordinate). Finally, we flatten the edge points and apply an MLP with two linear layers to obtain a 512-dimensional value embedding. In addition, we use a learnable special token \texttt{[EOS]} to indicate the end of the sequence.

\smallskip
\noindent {\bf Output sequence and embeddings.} As we cast constraint detection as a sequence generation problem, the relevant constraints are represented as a sequence of edges: $g(e_{j_1}) = (e_{j_1}, \ldots, e_{j_s})$. Edges in the output sequence are ordered according to their indices. Take Figure~\ref{fig:wireframe} as an example, the desired output sequence when using $e_1$ as the query edge would be $\splitatcommas{(e_1, e_6, e_{11}, e_{14}, e_{17}, e_{18}, \texttt{[EOS]})}$ for parallel constraint prediction, and $\splitatcommas{(e_1, e_2, e_3, e_4, e_5, e_7, e_8, e_9, e_{10}, e_{12}, e_{13}, e_{15}, e_{16}, \texttt{[EOS]})}$ for perpendicular constraint prediction. Same as the encoder, the input embeddings of the decoder consist of a value embedding and a position embedding. And we simply use an edge's contextual embedding from the encoder output as its value embedding.

\smallskip
\noindent {\bf Inference and filtering.} Since we predict a constraint sequence for each query edge, we may exploit the symmetry property of the constraints to filter out inconsistent predictions. Specifically, let $g(e_i)$ denote the edge prediction from query $e_i$, the final set of constraints should be:
$\{ (e_i, e_j) \mid e_i \in g(e_j), e_j \in g(e_i); \forall e_i \in \E, \forall e_j \in \E, i \neq j \}$. 

\section{Initial Value Prediction}
\label{sec:depth}

We now turn our attention to the estimation of the initial value of $\Z$. Similar to constraint detection, the number of geometric entities involved (\ie, vertices) is different for each line drawing. Thus, we again treat the input as a sequence and employ a Transformer-based architecture for sequence modeling. But unlike constraint detection, the number of outputs (\ie, depth values) is always equal to the number of vertices. Therefore, we may simplify the network architecture and use a Transformer decoder only to deal with the equal number of inputs and outputs.

\begin{figure}[t]
    \centering
    \includegraphics[width=0.95\linewidth]{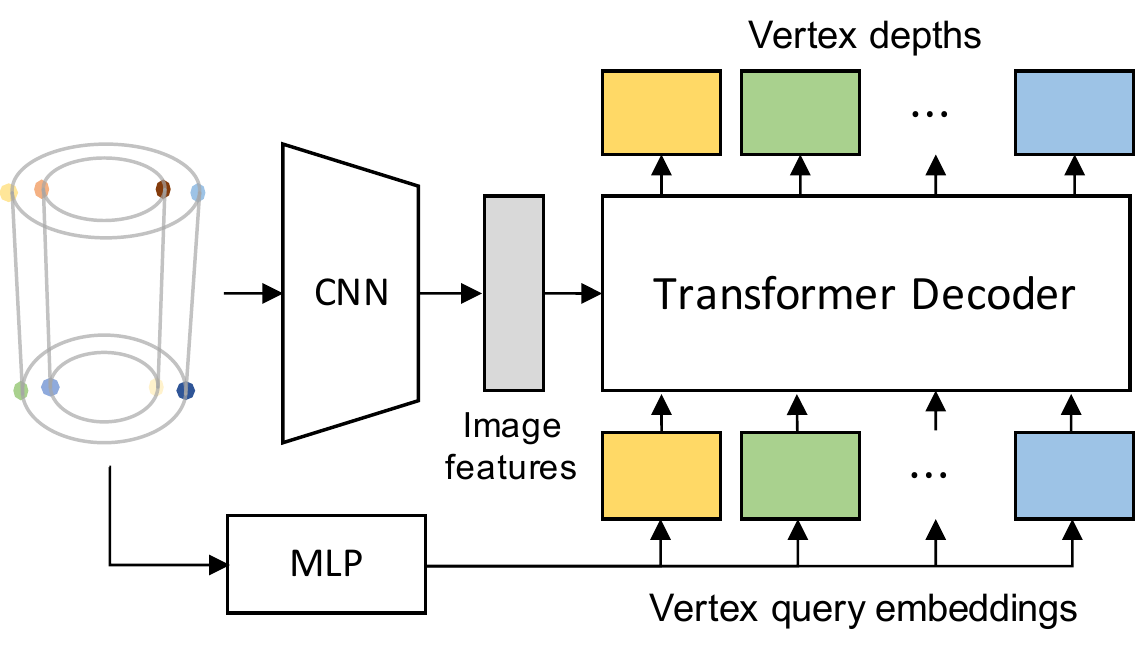}
    \caption{Depth estimation network.}
    \label{fig:depth}
    \vspace{-5mm}
\end{figure}

But it is impractical to infer depth values from the vertex coordinates only, because it ignores contextual information in the 2D line drawing such as the shape of edges and the topology of the object. To leverage the visual information, we convert the line drawing to a bitmap image, and use a convolutional neural network (CNN) to extract visual features. Specifically, we use a ResNet-18 backbone which takes input images of size $256\times 256$ and outputs feature maps of size $16\times 16\times d$, where $d$ is the dimensionality of the features. Finally, we flatten the features along the spatial dimensions to create a feature sequence of size $256\times d$.

Our overall network design is shown in Figure~\ref{fig:depth}. The Transformer decoder takes flattened visual features and vertex embeddings as input and outputs the depth value for each vertex. Same as edge embedding in the constraint detection network, we also use two embeddings for each vertex, \ie, value embedding and position embedding. For value embedding, we use an MLP with two linear layers to obtain a 512-dimensional embedding. We use scale-invariant loss~\cite{EigenPF14} to supervise the training of the depth estimation network.

\section{Experiments}
\label{sec:experiments}

\subsection{Experimental Setup}

\noindent {\bf Dataset.} Since no prior work has attempted to integrate deep learning with nonlinear optimization for 3D reconstruction, we build a benchmark for this novel task using a subset of CAD models from ABC dataset~\cite{KochMJWABAZP19}. We obtain the input line drawings from CAD models via pythonOCC\footnote{\url{https://github.com/tpaviot/pythonocc}}, a Python 3D development framework built upon the Open CASCADE Technology~\cite{occt}. To this end, we normalize the shape such that the half diagonal length of the bounding box is equal to 1, and place the camera at a fixed point with a distance of 6 from the object center, pointing towards the object. The camera focal length is set to 5.

In practice, we limit the edge type to line segments and circular arcs, but our methods are also applicable to other types. As ABC dataset~\cite{KochMJWABAZP19} contains many duplicated shapes, we filter the duplications based on the shape's topology and three orthogonal views. 
The remaining data is split into 52248 objects for training, 2903 objects for validation, and 2902 objects for testing.

\smallskip
\noindent {\bf Implementation details.} We implement our models with PyTorch and PyTorch Lightning. We use Adam optimizer~\cite{KingmaB15} with a learning rate of $5 \times 10^{-4}$. The batch size is 8 for constraint detection, and 256 for depth prediction. We train the constraint detection network for 200K iterations, and the depth estimation network for 120K iterations, on four NVIDIA RTX 3090 GPU devices.

\subsection{Experiments on Constraint Detection}
\label{sec:exp:constraint}

In this section, we conduct experiments to evaluate our constraint detection model. For geometric constraints, we consider the parallel constraints between two line segments (line-line), and perpendicular constraints between two line segments (line-line), or one line segment and one arc (line-arc).

To evaluate a constraint detection method, we use three standard metrics including precision, recall, and F1 score. Specifically, we regard each constraint as a pair of edges. A constraint prediction is correct if and only if the pair of edges matches a ground truth. %

\smallskip
\noindent{\bf Methods for comparison.} In this experiment, we compare our method to three widely recognized baseline methods, as described below. Note that detecting constraints which involve curved edges (\ie, arcs) poses additional challenge for existing methods, because the ``direction'' of an arc, on which these methods rely, is not well-defined. To resolve this issue, we replace each arc with a line segment connecting its two endpoints, based on the observations that (i) if a line is perpendicular to an arc, it is also perpendicular to its chord (the reverse is not necessarily true), and (ii) since our goal is to recover the depths of the vertices, it suffices to specify the constraints w.r.t. line segments connecting them. This way, technique developed for straight lines can be directly applied (in contrast, our method is not restricted to any specific edge types).

\begin{figure}[t]
    \centering
    \includegraphics[width=\linewidth]{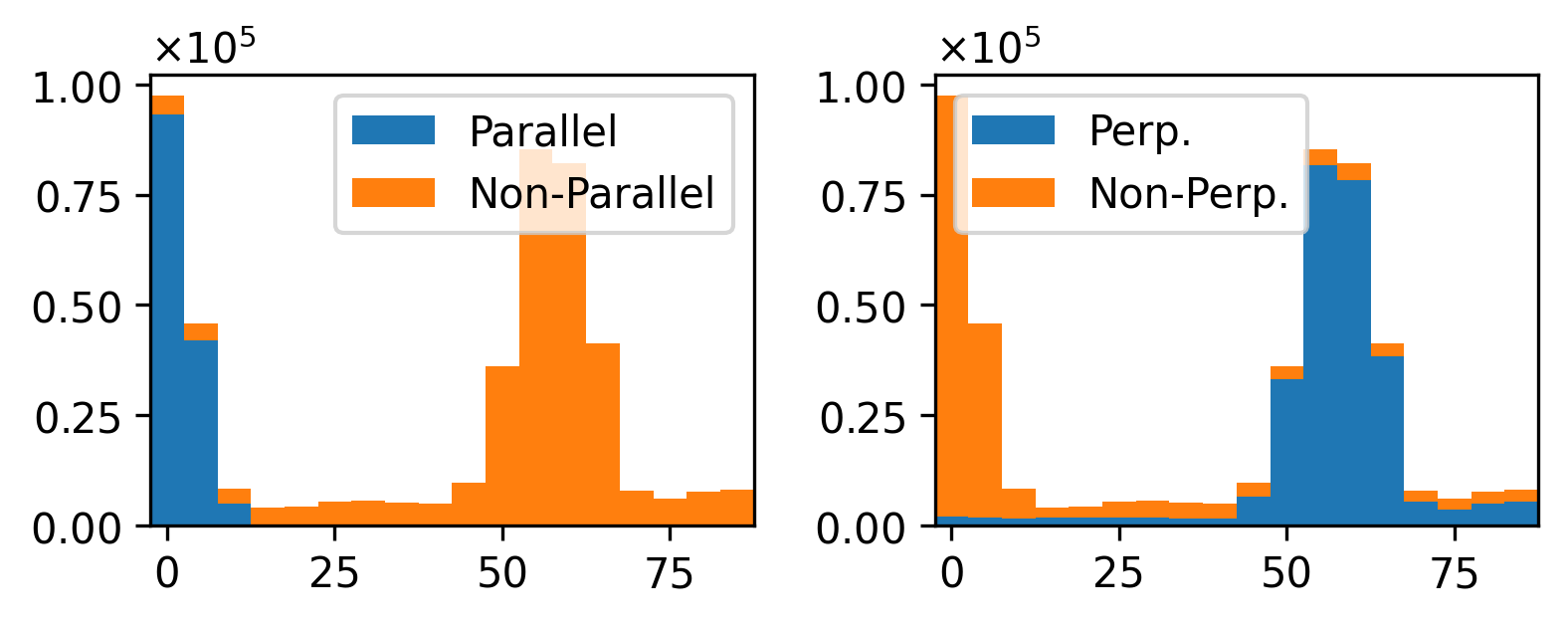}
    \vspace{-5mm}
    \caption{Angle distribution between lines in our test set.}
    \label{fig:dist}
    \vspace{-5mm}
\end{figure}

\underline{\em Heuristic rules:}  In prior work, simple heuristics were commonly used to detect geometric relationships. For example, in~\cite{LipsonS96,ZouPLCFL15,ZouL07}, two lines are considered to be parallel if their angle satisfies $\theta\leq 7^{\circ}$. To apply such heuristics, we tune the angle threshold on our test set. Figure~\ref{fig:dist} shows the angle distributions between lines in our test set. Accordingly, we set the condition to $\theta \leq 15^{\circ}$ for parallel relationship, and $\theta > 20^{\circ}$ for perpendicular relationship.

\underline{\em J-Linkage:} An alternative way to detect parallel lines is through vanishing point (VP) detection. Under the pinhole camera model, parallel lines in 3D space project to converging lines in the 2D drawing. The common point of intersection is called the \emph{vanishing point} (VP). 
In this work, we adopt J-Linkage~\cite{Tardif09}, a popular clustering-based approach for VP detection. For each VP, the associated lines are considered to be parallel to each other. 

With the VPs, perpendicular constraints may also be obtained by identifying the three main orthogonal directions (also known as the Manhattan directions~\cite{CoughlanY99}) in the 2D drawing. To this end, J-Linkage further selects three VPs which correspond to the three main directions in the 3D space. Any pair of lines that belong to two different main directions forms a perpendicular relationship. 

Obviously, some limitations exist for this method: it cannot detect perpendicular relationships between lines which do not align with the three main directions, or in an object which does not satisfy the Manhattan world assumption. 

\underline{\em True2Form:} As one of the state-of-the-art methods in sketch-based reconstruction, True2Form~\cite{XuCSBMS14} uses an initial baseline reconstruction to iteratively select applicable constraints, which are in turn used to refine the reconstruction. But to obtain the initial reconstruction, it relies on additional inputs in raw sketches, namely the smooth-crossings. Such information is not available in our problem. To remedy this issue, we use the outputs from our proposed initial depth estimation network (Section~\ref{sec:depth}) instead.

Following~\cite{XuCSBMS14}, we use at most four iterations to select one type of constraints. In each iteration, we minimize $\sum_i (Z_i - Z_i^0)^2 + w_c \sum_k L(\alpha_k) C_k^2$, where $Z_i^0$ is the initial depth of $i$-th vertex estimated by our network, $C_k$ is the $k$-th constraint to be considered, $L(\alpha_k)$ is an angle-based likelihood function, and $w_c$ is a fixed weight. All parameters are set to the default values as suggested in~\cite{XuCSBMS14}.

\begin{table}[t]
    \centering
    \setlength{\tabcolsep}{2pt}
    \begin{tabular}{l|c|cccc}
        \toprule
        Type & methods & F1 & precision & recall & time (s) \tabularnewline
        \midrule
        \multirow{4}{*}{Parallel} & Heuristic rules & 96.51 & 94.68 & {\bf 99.59} & 0.004 \tabularnewline
        & J-Linkage~\cite{Tardif09} & 94.50 & 96.35 & 90.80 & 1.546 \tabularnewline
        & True2Form~\cite{XuCSBMS14} & 92.52 & {\bf 99.18} & 89.22 & 0.273 \tabularnewline
        & Ours & {\bf 99.08} & 99.11 & 99.20 & 0.054 \tabularnewline
        \midrule
        \multirow{4}{*}{Perp.} & Heuristic rules & 87.83 & 81.77 & {\bf 97.57} & 0.004 \tabularnewline
        & J-Linkage~\cite{Tardif09} & 76.73 & 87.70 & 71.64 &1.955 \tabularnewline
        & Ture2Form~\cite{XuCSBMS14} & 94.38 & 97.94 & 90.07 & 1.474 \tabularnewline
        & Ours & {\bf 97.24} & {\bf 98.82} & 96.07 & 0.106 \tabularnewline
        \bottomrule
    \end{tabular}
    \caption{Comparison with prior work on constraint detection.}
    \vspace{-3mm}
    \label{tab:prior}
\end{table}

\smallskip
\noindent{\bf Quantitative results.}  As shown in Table~\ref{tab:prior}, for parallel constraint detection, our method achieves a nearly perfect result. Meanwhile, using simple heuristic rules still leads to a substantial number of false positives on this seemingly straightforward task. Leveraging an initial reconstruction, Ture2Form also achieves very high precision. But its recall is not as high, partly due to errors and noises in the initial depth values predicted by our network.

For perpendicular constraint detection, the performance of heuristic rules and J-Linkage further degrades. Notably, J-Linkage has a low recall ($71.64\%$) because, as we mentioned before, it cannot detect perpendicular constraints involving edges which do not align with the three main directions. In comparison, our method achieves high percentages in the precision and recall metrics, as well as the highest F1 score. 

\smallskip
\noindent{\bf Is deep learning really needed for constraint detection?} We observe in Figure~\ref{tab:prior} that True2Form and our method both have very high precisions. This is critical for geometric constraint solving because even a single incorrect constraint could lead to drastically different solutions. In contrast, achieving a high recall is not as important because, in practice, the number of constraints is usually larger than the number of variables. Thus, one may wonder if it is necessary to use deep learning for constraint detection.

We argue that the benefit of our constraint detection method is two-fold. \emph{First}, it disentangles constraint detection from initial value prediction. Recall that, True2Form relies on a baseline reconstruction generated by our method to detect constraints. In contrast, our deep constraint detection network works independently, thus can find constraints when a baseline reconstruction is erroneous or unavailable. \emph{Second}, it has a lower time complexity compared to the iterative optimization scheme employed by True2Form. As shown in~Figure~\ref{tab:prior}, the computational time for True2Form grows quickly as more constraints are considered.

\subsection{Experiments on 3D Reconstruction}
\label{sec:exp:3d}

\begin{table*}[t]
    \centering
    \begin{tabular}{l|l|cccccc}
        \toprule
        Constraint & \#Iter & Success & Unsolvable & Fail & Wrong (I) & Wrong (II) & Wrong (III) \tabularnewline
        \midrule
        \multirow{2}{*}{GT} & 1 & 2408 ($83.0\%$) & 151 & 74 & 0 & 269 & 0 \tabularnewline
        & 10 & 2626 ($90.5\%$) & 29 & 11 & 0 & 157 & 79 \tabularnewline
        \midrule
        \multirow{2}{*}{Prediction} & 1 & 2252 ($77.6\%$) & 132 & 96 & 176 & 246 & 0 \tabularnewline
        & 10 & 2482 ($85.5\%$) & 35 & 6 & 138 & 138 & 103 \tabularnewline
        \bottomrule
    \end{tabular}
    \caption{Performance of our method on geometric constraint solving.}
    \label{tab:solving}
    \vspace{-5mm}
\end{table*}

Next, we verify the effectiveness of using constraints and initial values predicted by our deep models in optimization-based 3D reconstruction. To evaluate the reconstruction results, we compute the absolute depth difference of the vertices between the prediction and ground truth. A shape is successfully reconstructed only if the differences for \emph{all} vertices are smaller than a certain threshold. In all experiments, we set the threshold to $10^{-3}$. 

To disentangle the influence of constraint detection and that of initial value prediction, we run experiments with both the ground truth constraints and the predicted constraints, and report the results in Table~\ref{tab:solving}. 
Also, as discussed in Section~\ref{sec:selecting_constraint}, the numerical method (\ie, \texttt{fsolve}) may fail to obtain a solution depending on the initial value and the selected constraints. To this end, we further report the results by running our pipeline for $N=1$ and $N=10$ times, respectively. Comparing the results in Table~\ref{tab:solving}, we can see that repeating the process of selecting and solving constraints for 10 times can mostly eliminate this issue and improve the success rate (\eg, from $83.0\%$ to $90.5\%$ for GT constraints).

Now, to better understand the mistakes made by our method (with $N=10$), we classify the failure cases into five categories, as we explain below:

\begin{wrapfigure}{r}{0.5\linewidth}
    \centering
    \vspace{-5mm}
    \setlength{\tabcolsep}{3pt}
    \begin{tabular}{cc}
        \includegraphics[width=0.45\linewidth]{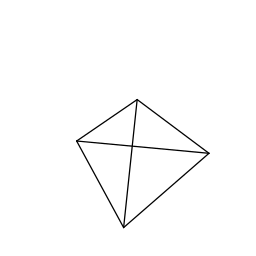} &
        \includegraphics[width=0.45\linewidth]{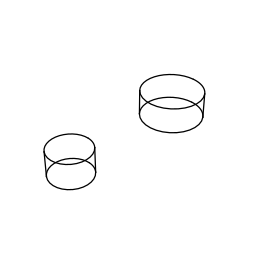}
    \end{tabular}
    \caption{Two examples of under-constrained objects in our dataset.}
    \label{fig:unsolvable}
    \vspace{-5mm}
\end{wrapfigure}

\underline{\em Unsolvable}: Cases where it fails to find a sufficient set of constraints. This typically occurs when the object is under-constrained. We show two such examples in Figure~\ref{fig:unsolvable}. The first case is a tetrahedron where no meaningful parallel, perpendicular, or face planarity constraint exists. The second case consists of multiple disjoint parts and there is no constraint available to recover the relative scale between the parts.

\underline{\em Fail}: Cases where a sufficient set of constraints can be found but \texttt{fsolve} fails to find a solution. This is mainly due to stability issues of the numerical method. Such cases are rare with $N=10$.

\underline{\em Wrong (I)}: Cases where it fails to find a sufficient set of \emph{correct} constraints in all ten attempts. This only occurs when predicted constraints are used. And for each of these cases, the performance of constraint detection model is particularly poor. As a result, at least one incorrect constraint is selected in every run.

\underline{\em Wrong (II)}: Cases where it finds a sufficient set of correct constraints in at least one attempt, but fails to obtain the correct 3D model. This typically occurs when the predicted initial value of $\Z$ deviates too much from the ground truth.

\underline{\em Wrong (III)}: Cases where it finds the correct solution in at least one attempt, but chooses an incorrect solution as the final output. A close look at these cases reveals that such mistakes are mainly due to the strict tolerance ($10^{-3}$) we use in the experiments. In these cases, the desired shape is found, but the precision of the numerical solution is not high enough (\ie, $>10^{-3}$).

\begin{figure}[t]
    \centering
    \includegraphics[width=0.8\linewidth]{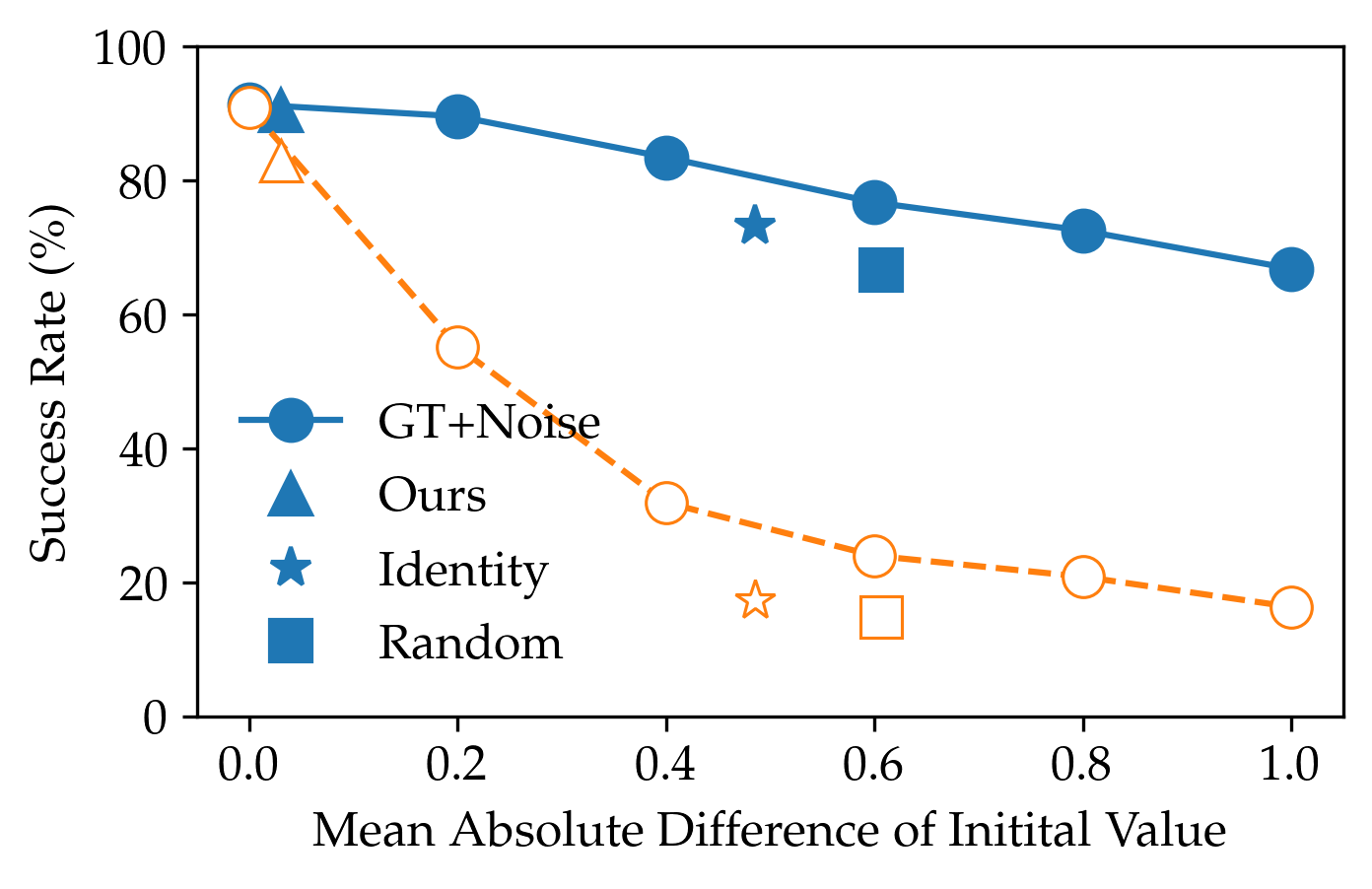}\\
    \includegraphics[width=0.8\linewidth]{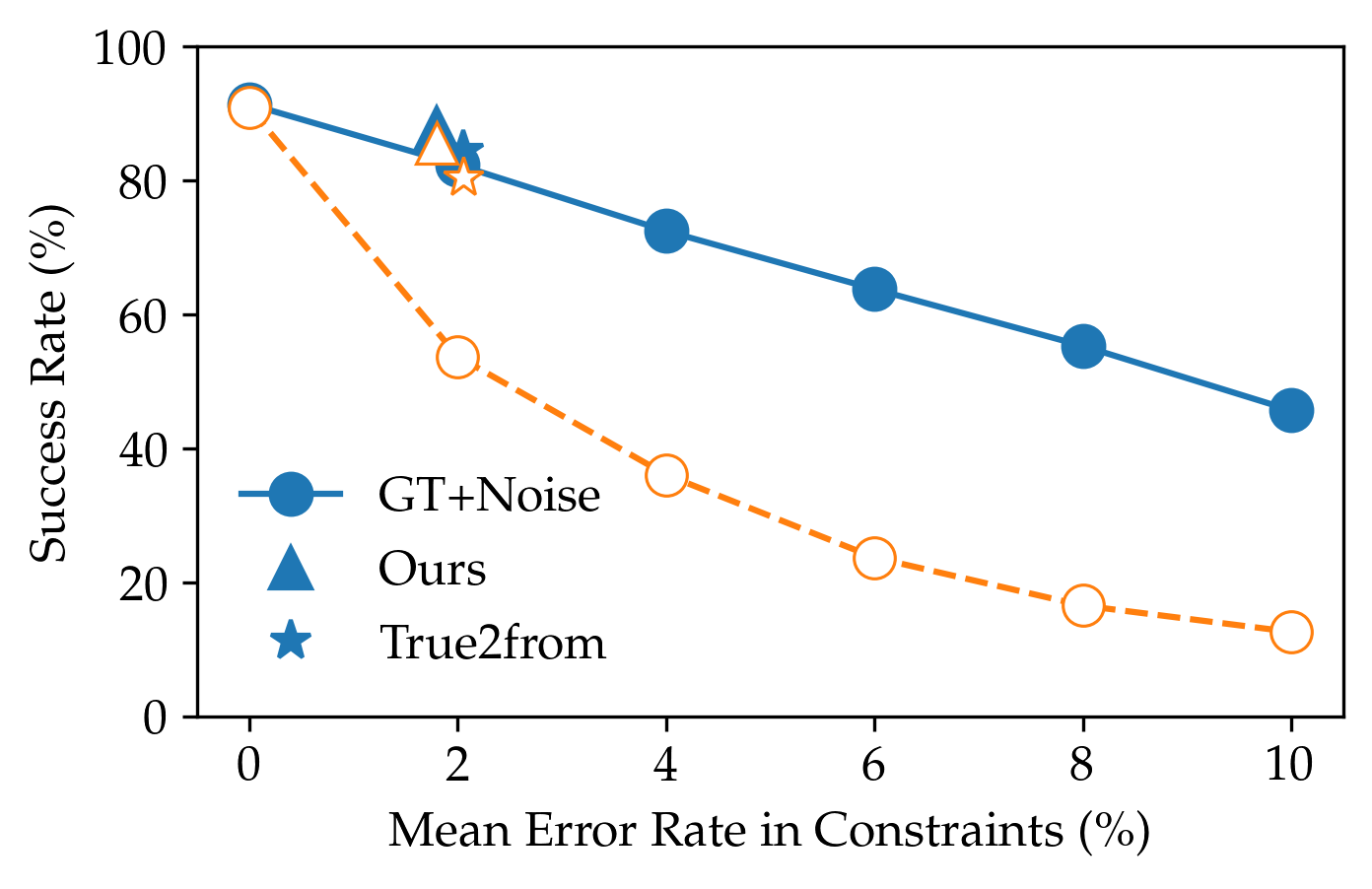}
    \caption{Influences of initial value prediction and constraint detection. We plot results for $N=1$ (in orange) and $N=10$ (in blue). {\bf Top}: Success rate w.r.t. different initialization strategies. {\bf Bottom}: Success rate w.r.t. different constraint detection methods.}
    \label{fig:ablation}
    \vspace{-4mm}
\end{figure}

\begin{figure*}[t]
    \centering
    \small
    \setlength{\tabcolsep}{5pt}
    \begin{tabular}{c|cc|c||c|cc|c}
        Input & \multicolumn{2}{c|}{Ours} & AtlasNet & Input & \multicolumn{2}{c|}{Ours} & AtlasNet \tabularnewline
        \midrule

        \includegraphics[width=0.1\linewidth]{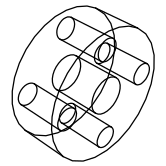} &
        \includegraphics[width=0.1\linewidth]{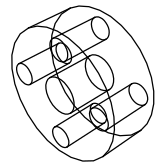} &
        \includegraphics[width=0.1\linewidth]{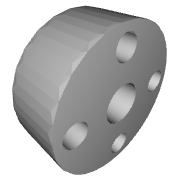} &
        \includegraphics[width=0.1\linewidth]{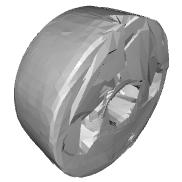} &
        \includegraphics[width=0.1\linewidth]{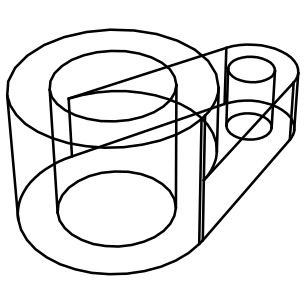} &
        \includegraphics[width=0.1\linewidth]{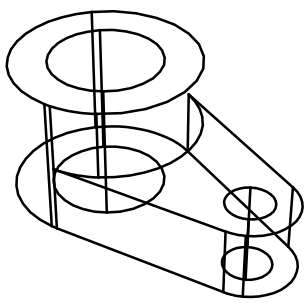} &
        \includegraphics[width=0.1\linewidth]{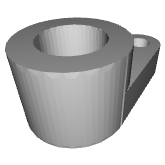} &
        \includegraphics[width=0.1\linewidth]{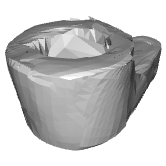}
        \tabularnewline

        \includegraphics[width=0.1\linewidth]{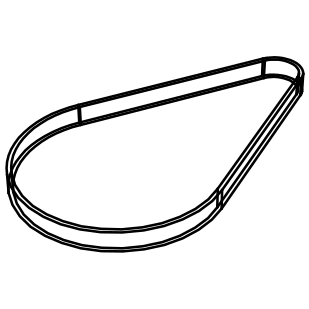} &
        \includegraphics[width=0.1\linewidth]{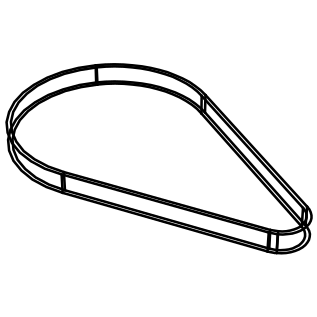} &
        \includegraphics[width=0.1\linewidth]{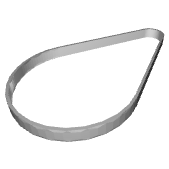} &
        \includegraphics[width=0.1\linewidth]{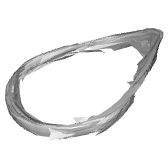} &
        \includegraphics[width=0.1\linewidth]{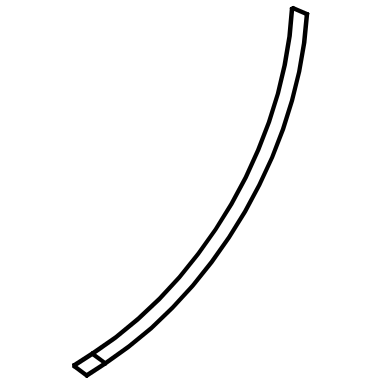} &
        \includegraphics[width=0.1\linewidth]{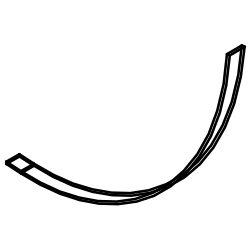} &
        \includegraphics[width=0.1\linewidth]{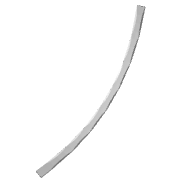} &
        \includegraphics[width=0.1\linewidth]{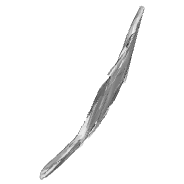}
        \tabularnewline

        \includegraphics[width=0.1\linewidth]{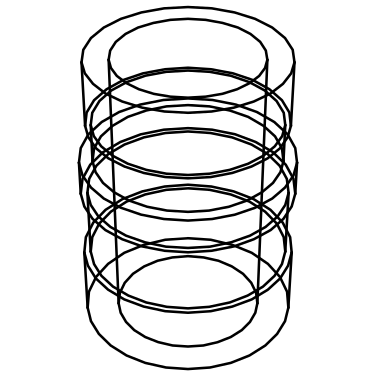} &
        \includegraphics[width=0.1\linewidth]{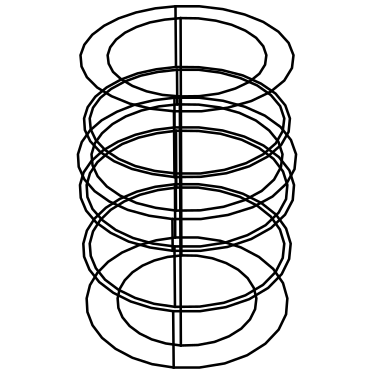} &
        \includegraphics[width=0.1\linewidth]{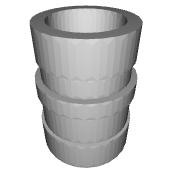} &
        \includegraphics[width=0.1\linewidth]{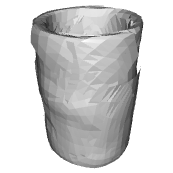} &
        \includegraphics[width=0.1\linewidth]{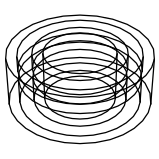} &
        \includegraphics[width=0.1\linewidth]{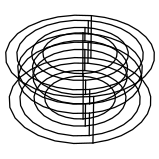} &
        \includegraphics[width=0.1\linewidth]{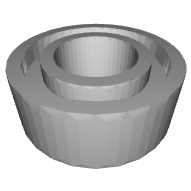} &
        \includegraphics[width=0.1\linewidth]{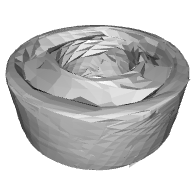}
        \tabularnewline

        \includegraphics[width=0.1\linewidth]{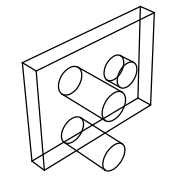} &
        \includegraphics[width=0.1\linewidth]{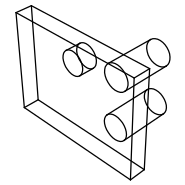} &
        \includegraphics[width=0.1\linewidth]{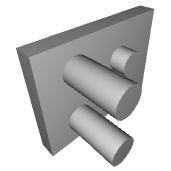} &
        \includegraphics[width=0.1\linewidth]{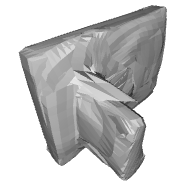} &
        \includegraphics[width=0.1\linewidth]{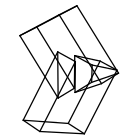} &
        \includegraphics[width=0.1\linewidth]{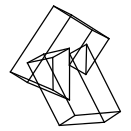} &
        \includegraphics[width=0.1\linewidth]{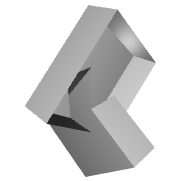} &
        \includegraphics[width=0.1\linewidth]{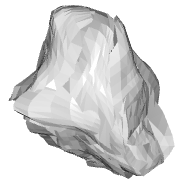}
        \tabularnewline

        \includegraphics[width=0.1\linewidth]{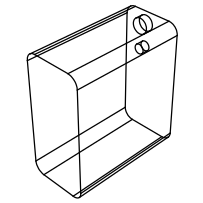} &
        \includegraphics[width=0.1\linewidth]{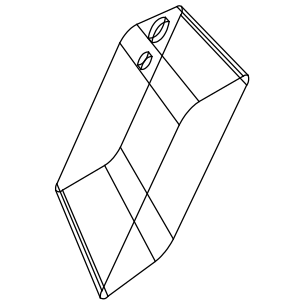} &
        \includegraphics[width=0.1\linewidth]{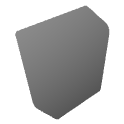} &
        \includegraphics[width=0.1\linewidth]{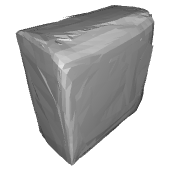} &
        \includegraphics[width=0.1\linewidth]{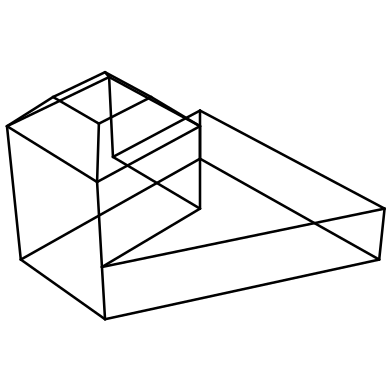} &
        \includegraphics[width=0.1\linewidth]{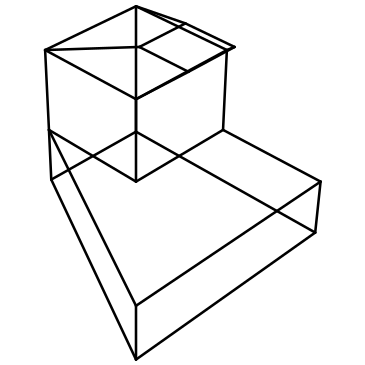} &
        \includegraphics[width=0.1\linewidth]{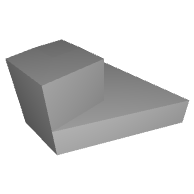} &
        \includegraphics[width=0.1\linewidth]{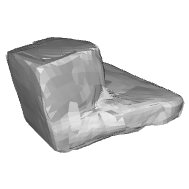}
        \tabularnewline

        \includegraphics[width=0.1\linewidth]{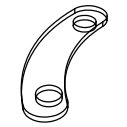} &
        \includegraphics[width=0.1\linewidth]{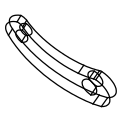} &
        \includegraphics[width=0.1\linewidth]{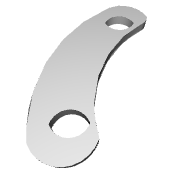} &
        \includegraphics[width=0.1\linewidth]{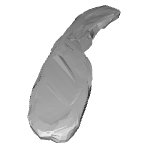} &
        \includegraphics[width=0.1\linewidth]{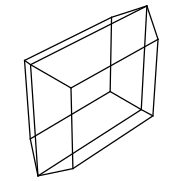} &
        \includegraphics[width=0.1\linewidth]{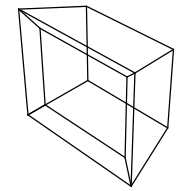} &
        \includegraphics[width=0.1\linewidth]{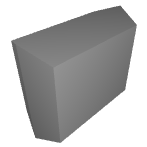} &
        \includegraphics[width=0.1\linewidth]{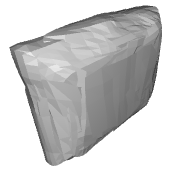}
        \tabularnewline

    \end{tabular}
    \caption{3D reconstruction results. For our method, we show the reconstructed 3D wireframe (from a different viewpoint) and the mesh model.}
    \label{fig:mesh}
    \vspace{-5mm}
\end{figure*}

\smallskip
\noindent {\bf Influence of initial value prediction.} From the above analysis we can see that, initial value prediction mostly contributes to failure cases in the \emph{Wrong (II)} category, which constitutes $138/2902=4.8\%$ of the testing set. 

To see the significance of this result, we compare the success rate using the predicted initial value with those of two popular strategies: (i) \emph{Identity}: initializing with a flat 2D projection by setting the depths of all vertices to the true value of the first vertex, and (ii) \emph{Random}: initializing by randomly sampling depth in the interval $[5,7]$ for each vertex. As a reference, we also generate initial values by systematically perturbing the ground truth with different levels of noises, and plot the corresponding success rate curve (\emph{GT+noise}). As shown in Figure~\ref{fig:ablation} (top), our method finds initial values with a much lower mean absolute difference (0.03) than \emph{Identity} (0.49) or \emph{Random} (0.61), resulting in a success rate which is very close to that of using ground truth.

\smallskip
\noindent {\bf Influence of constraint detection.} In the meantime, errors in the detected constraints mostly contribute to failure cases in the \emph{Wrong (I)} category, which constitutes $138/2902=4.8\%$ of the testing set. 

We also compare our method with an alternative approach that employs True2Form for parallel and perpendicular constraint detection, as described in Section~\ref{sec:exp:constraint}. As shown in Figure~\ref{fig:ablation} (bottom), this approach performs slightly worse than our method. This is expected as True2Form has slightly worse scores on constraint detection (Table~\ref{tab:prior}). More importantly, it suggests that it is possible to obtain reliable constraint detection results via deep learning, without relying on a baseline reconstruction or an iterative selection scheme (as True2Form does).

\smallskip
\noindent {\bf Comparison with end-to-end 3D reconstruction method.} Most deep learning methods for 3D reconstruction generate unstructured point clouds or meshes as output. In this experiment, we compare our pipeline with AtlasNet~\cite{GroueixFKRA18}, a leading deep learning method for single-view 3D reconstruction\footnote{\url{https://github.com/ThibaultGROUEIX/AtlasNet}}.

Figure~\ref{fig:mesh} shows the 3D models obtained by the two methods. In the first three rows, we show several examples for which our optimization-based method reconstructs the correct models. For example, our approach is able to reconstruct the thin models in the \emph{second row} and the complex models in the \emph{third row}. In comparison, although AtlasNet can produce a coarse 3D model for each case, it struggles to recover the correct object topology (\eg, holes) and fine details. 

In the last three rows of Figure~\ref{fig:mesh}, we show some failed cases of our method. In the \emph{fourth row}, we show two cases with imperfect face predictions. In the first case, the face identification method fails to recover the correct topology of the face with a hole. In the second case, the reconstructed model is incomplete due to missed faces. In the \emph{fifth row}, we show two cases where (partially) flat models are recovered due to errors in constraint detection. In the \emph{last row}, a sufficient set of constraint is found for each case, but the optimization finds a solution that is different from the ground truth (probably due to initial value prediction).

\begin{figure}[t]
    \centering
    \includegraphics[width=0.9\linewidth]{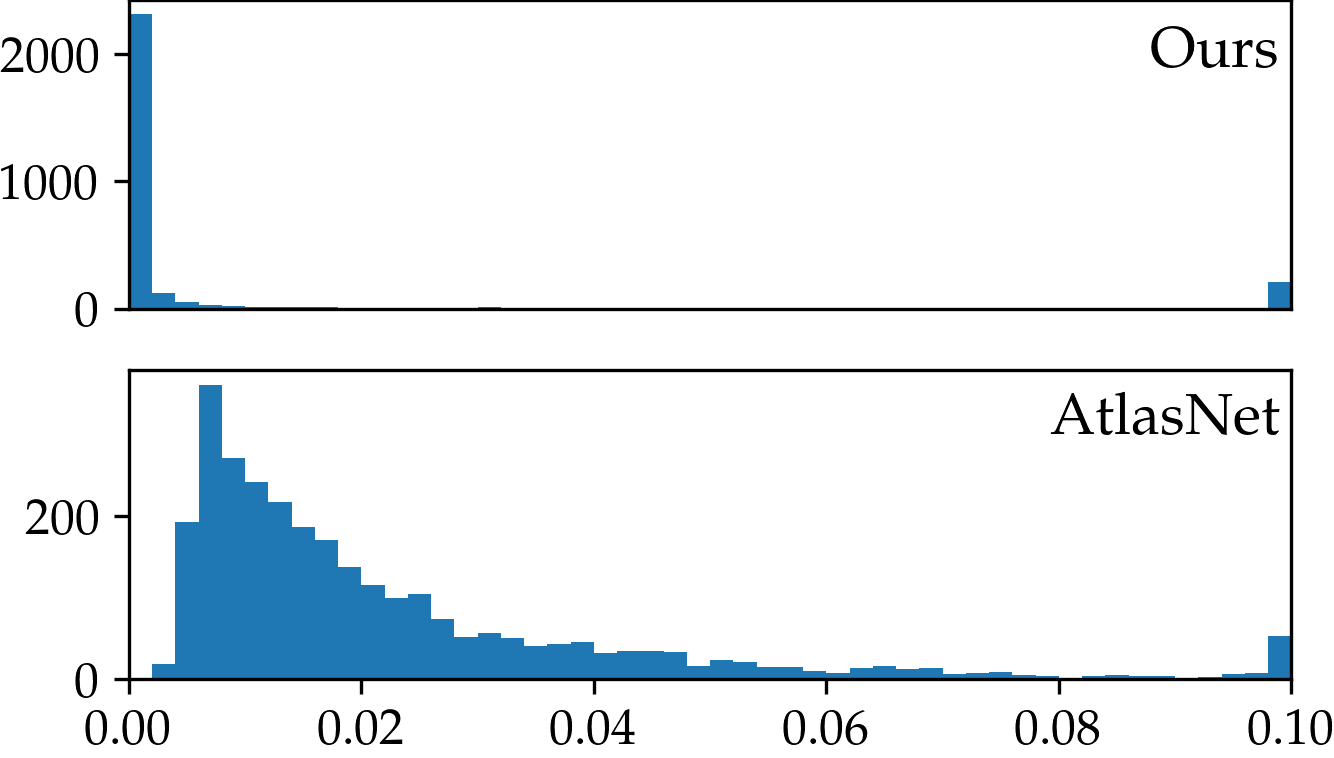}
    \caption{Histogram of Chamfer distances between the predicted model and the ground truth.}
    \label{fig:chafmer}
    \vspace{-5mm}
\end{figure}

Finally, for each object in the test set, we compute the Chamfer distance between the predicted model and the ground truth, and plot the histogram in Figure~\ref{fig:chafmer}. As one can see, the 3D models obtained by our method are more accurate.

\section{Conclusion}

We have proposed a novel scheme for 3D reconstruction from single line drawings which integrates deep learning and nonlinear optimization for geometric constraint solving, and demonstrated its effectiveness on a large-scale CAD dataset by comparing with the state of the arts. Significant improvement to the success rate of 3D reconstruction have been attained.

A notable limitation of this work is that the current pipeline assumes the input line drawing is clean and noise-free. To handle real-world data such as hand-drawn sketches and scanned images, additional steps may be required to clean up the data. In the future, we plan to further integrate these steps into our pipeline to build a complete system for sketch-based 3D modeling. 

\section*{Acknowledgements}

This work was supported in part by the Key R\&D Program of Zhejiang Province (No.~2022C01025), the National Natural Science Foundation of China (No.~62102355), and the Natural Science Foundation of Zhejiang Province (No.~LQ22F020012).

{\small
\bibliographystyle{ieee_fullname}
\bibliography{constraint}
}

\end{document}